# Evolution of Digital Logic Functionality Via a Genetic Algorithm

**Christopher M. Frenz**[1*], **Steve Peters**[1], **and Wilson Julien**[1]
[1]Department of Computer Engineering Technology
New York City College of Technology (CUNY)
Brooklyn, NY 11201
[*]Corresponding Author: cfrenz@citytech.cuny.edu

**Abstract** - *Digital logic forms the functional basics of most modern electronic equipment and as such the creation of novel digital logic circuits is an active area of computer engineering research. This study demonstrates that genetic algorithms can be used to evolve functionally useful sets of logic gate interconnections to create useful digital logic circuits. The efficacy of this approach is illustrated via the evolution of AND, OR, XOR, NOR, and XNOR functionality from sets of NAND gates, thereby illustrating that evolutionary methods have the potential be applied to the design of digital electronics.*

## 1   Introduction

Digital logic forms the cornerstone of most modern electronic equipment, and as such the creation of functionally useful digital logic circuits is a key area of research for the computer industry. Digital logic functionality is generally comprised of a large number of interconnected logic gates, where the order and arrangements of such interconnections play a critical role in achieving the appropriate logic functionality. The significance of these interconnections is probably most readily demonstrated by the fact that appropriately connected combinations of NAND gates can be used to create the functionality of any other type of logic gate [1].

Determining the appropriate interconnections of such logic gates can be viewed as a search problem in that an appropriate set of interconnections must be chosen from all possible sets of interconnections in order to yield the desired functionality. Genetic algorithms often provide a useful way for handling such search problems in that over time a solution to many search and optimization problems can be evolved [2]. As such, this study seeks to determine the feasibility of applying a genetic algorithm to the determination of NAND gate interconnections required to produce digital logic functionality.

## 2   Methods

### 2.1   The Genetic Algorithm

The basis for this genetic algorithm is that each NAND gate consists of two genes, where a gene is defined as an input for that particular gate, such that a circuit consisting of three NAND gates would have six total genes. The potential alleles for each gene are comprised of any of the possible inputs for the gate, such as input into the circuit from an external source or input consisting of an output from another gate within the circuit. The algorithm initially generates a population of NAND gate circuits of user specified size and randomly assigns the allele values which control gate interconnections within the members of the population. A fitness function is then applied to each member of the population, where fitness is computed by ascertaining how close the circuit comes to producing the desired outputs from the external inputs provided. As such, a member with a perfect fitness would be the desired logic circuit, since it would be able to take the provided inputs and turn them into the desired outputs for all potential input values, and when a member of the population reaches this fitness level the results are output and the program that employs the algorithm is terminated. A circuit with zero fitness would not be able to generate any outputs correctly for the inputs provided, and all other levels of fitness would be intermediary to these two extremes. Upon application of the fitness function all circuits with zero fitness are removed from the population and all remaining members are given equal opportunity to breed, since it is assumed that their partial fitness is indicative that they may have a partially valid circuit design. For purposes of breeding, two members of the population are selected and for each gene there is a 45% chance that the child's gene will come from one parent, a 45% that the gene will come from the other parent, and a 10% of the child's gene arising from mutation to another valid input source (Figure 1). The purpose of this mutation is to prevent a population from becoming too uniform around a given set of genes. The size of the population is fixed so that for each generation the population consists of the same number of members as the initial population. After breeding occurs the population is once again exposed to the fitness function and the process repeated until a digital circuit capable of producing the correct outputs for all input values is generated. Code for the algorithm was written in Perl and all execution of the code was performed on an Intel Pentium 4 processor.

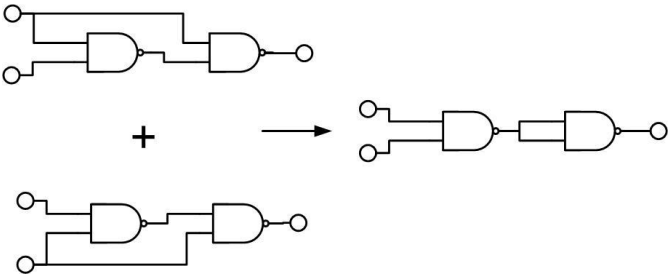

Figure 1: Demonstration of how two sets of NAND gates that are partially correct could combine during breeding to create a fully functional AND gate.

### 2.2 Testing the Algorithm

The algorithm was used to generate 2 input AND, OR, NOR, XOR, and XNOR gate functionality from a digital circuit comprised of NAND gates. In all cases the external inputs provided consisted of all possible two input binary combinations (e.g. 0 and 0, 0 and 1, 1 and 0, and 1 and 1) and the circuit output was output that would be expected for the gate being generated. For all design attempts a small population size of 10 was used to limit the chance of the random generation of population members consistently generating a solution by chance and thus allowing the evolutionary aspects of the algorithm to actually generate the desired functionality. All design attempts were run at least 10 times and the reported numbers of generations required to create the desired functionality from a population of 10 members are the averages of these attempts.

### 2.3 Plots and Figures

Plots were generated in GraphPad Prism version 4.01 and circuit diagrams were prepared using Microsoft Visio Professional 2007.

## 3 Results and Discussion

The genetic algorithm detailed above was able to successfully evolve NAND gate interconnections to produce digital logic functions corresponding to the logic operations of AND, OR, NOR, XOR, and XNOR, and a representative set of interconnections is demonstrated for the XOR functionality (Figure 2). Given that these logic functions form the basic building blocks for more sophisticated computational functionality such as binary adders and binary multipliers the results are indicative that the algorithm has potential applications in the design of novel forms of computational circuitry, since creating more complex circuitry could be accomplished by expanding upon the numbers of gates that comprise each member of the population. The potential value for such algorithms will likely increase as the novelty and complexity of computational circuitry increases, such as in the inclusion of new instruction sets within a computer processor [3] or in the use of Field Programmable Gate Arrays to create specialized computational devices [4].

Moreover, for OR, NOR, and XNOR, multiple sets of working gate interconnections were actually evolved, indicating that the genetic algorithm is able to determine more that just a single solution to the problem. This has the added benefit in that the genetic algorithm could enable hardware developers to choose the most beneficial of the possible solutions, and not necessarily be limited to a single solution for yielding the desired functionality.

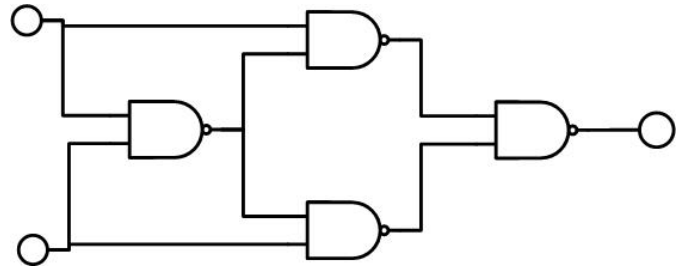

Figure 2: Representative set of NAND interconnections required to produce XOR functionality.

The average number of generations it took for the genetic algorithm to yield a solution to the desired logic functionality, for a population of 10 sets of gates, is demonstrated in Figure 3. It is notable that as the number of gates required to solve the problem increases, so too does the number of generations required find a solution (I.e. AND – 2 gates, OR- 3 gates, NOR and XOR – 4 gates, XNOR- 5 gates). One of the reasons for the high number of generations for the evolution XOR, NOR, and XNOR functionality is the small population size (10) used in these simulations. With an increase in population size the average number of generations required decreases, for example increasing the population size to 20 drops the average number of generations required to generate an XNOR gate to 130 generations, or less than half of the number of generations required for a population of 10. This is believed to result from the greater genetic diversity of the population and hence an increase in the likelihood of that circuits combine in novel ways during breeding. These findings are therefore suggestive that as the number of gates present in each member of the population increases, so to should the size of the population used in the simulation.

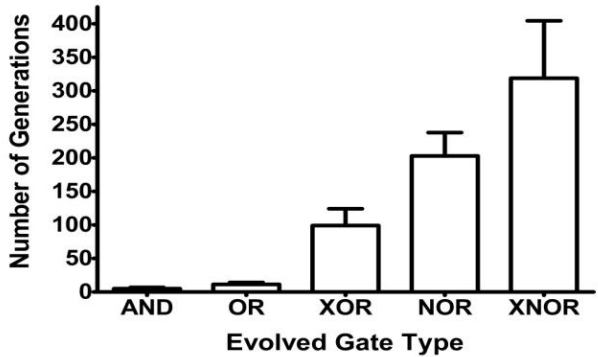

Figure 3: Average number of generations required to evolve each logic operation from a population comprised of 10 sets of NAND gates.

In all, these findings provide some preliminary proof that evolutionary methods are applicable to the design of digital logic circuits, however, the true utility of such approaches cannot be properly evaluated until attempts are made to apply such methods to the design of more sophisticated digital logic circuits.

## 4  Acknowledgements

This work was supported in part by the Louis Stokes Alliance for Minority Participation.